%% file: main.tex
\documentclass[letterpaper, 10 pt, conference]{ieeeconf}
\IEEEoverridecommandlockouts
\usepackage{cite}
\usepackage{amsmath,amssymb,amsfonts}
\usepackage{algorithmic}
\usepackage{graphicx}
\usepackage{textcomp}
\usepackage{xcolor}
\usepackage{subfigure}
\usepackage{subcaption}
\usepackage{xfrac}
\usepackage{fontawesome5}
\usepackage[switch]{lineno} 
\usepackage{caption} 
\usepackage{ifthen}
\usepackage{booktabs}
\def\BibTeX{{\rm B\kern-.05em{\sc i\kern-.025em b}\kern-.08em
    T\kern-.1667em\lower.7ex\hbox{E}\kern-.125emX}}
\begin{document}

\newcommand{\todo}[1]{\textcolor{red}{#1}}

\DeclareRobustCommand{\sgripith}[3]{%
  \ensuremath{%
    \ifthenelse{\equal{#3}{true}}%
      {\hat{s}_{#2}^{(#1)}}%
      {s_{#2}^{(#1)}}%
  }%
}
\newcommand{\stheta}{\ensuremath{s_t^\theta}}
\newcommand{\sthetahat}{\ensuremath{\hat{s}_t^\theta}}

\newcommand{\nonconform}{\faExclamationTriangle}

\newcommand{\wm}{\texttt{WM}}
\newcommand{\wmunc}{\wm\ uncertainty}
\newcommand{\wmprederr}{\wm\ prediction error}
\newcommand{\aenc}{\texttt{AE}}
\newcommand{\aencrecon}{\aenc\ reconstruction error}
\newcommand{\aenczsim}{\aenc\ sim({\ensuremath{z, z_{\textnormal{safe}}}})}
\newcommand{\logpzo}{\texttt{logpZO}}
\newcommand{\sparc}{\texttt{SPARC}}
\newcommand{\pcakmeans}{\texttt{PCA K-means}}
\newcommand{\random}{{Random}}

\newcommand{\datasetname}{Bimanual Cable Manipulation}

\newcommand{\manifold}{\ensuremath{\mathcal{M}_{\textnormal{nom}}}}
\newcommand{\realR}{\ensuremath{\mathbb{R}}}

\newcommand{\numnominalfit}{7}
\newcommand{\numnominaleval}{7}
\newcommand{\numfailure}{9}
\newcommand{\numtotal}{16} 

\newcommand{\symbolnominalfit}{\ensuremath{N_{\checkmark}}}

\newcommand{\symbolnominaleval}{\ensuremath{N_{\checkmark}}}

\newcommand{\symbolfailure}{\ensuremath{N_{\times}}}

\newcommand{\symboltotal}{\ensuremath{N}}

\title{\LARGE \bf{Foundational World Models Accurately \\Detect Bimanual Manipulator Failures
}}



\author{Isaac R. Ward$^{\dagger,*,1}$, Michelle Ho$^{*,1}$, Houjun Liu$^{1}$,  Aaron Feldman$^{1}$, Joseph Vincent$^{1}$, \\ Liam Kruse$^{1}$, Sean Cheong$^{2}$, Duncan Eddy$^{1}$, Mykel J. Kochenderfer$^{1}$ and Mac Schwager$^{1}$
\thanks{$^\dagger$Corresponding author.} 
\thanks{$^*$Authors contributed equally to this work.} 
\thanks{$^{1}$Stanford University, Stanford, California, 94305, USA. {\tt\small \{irward, mtho, houjun, lkruse, aofeldma, lkruse, deddy, schwager, mykel\}@stanford.edu}}
\thanks{$^{2}$Watney Robotics, San Francisco, California, 94110, USA. \tt\small sean@watneyrobotics.com}
\thanks{*Watney Robotics provided funds to support this work.}
\thanks{*Toyota Research Institute provided funds to support this work.}
}

\maketitle




\begin{abstract}
Deploying visuomotor robots at scale is challenging due to the potential for anomalous failures to degrade performance, cause damage, or endanger human life. Bimanual manipulators are no exception; these robots have vast state spaces comprised of high-dimensional images and proprioceptive signals. Explicitly defining failure modes within such state spaces is infeasible. In this work, we overcome these challenges by training a probabilistic, history informed, world model within the compressed latent space of a pretrained vision foundation model (NVIDIA's Cosmos Tokenizer). The model outputs uncertainty estimates alongside its predictions that serve as non-conformity scores within a conformal prediction framework. We use these scores to develop a runtime monitor, correlating periods of high uncertainty with anomalous failures. To test these methods, we use the simulated Push-T environment and the \datasetname\ dataset, the latter of which we introduce in this work. This new dataset features trajectories with multiple synchronized camera views, proprioceptive signals, and annotated failures from a challenging data center maintenance task. We benchmark our methods against baselines from the anomaly detection and out-of-distribution detection literature, and show that our approach considerably outperforms statistical techniques. Furthermore, we show that our approach requires approximately one twentieth of the trainable parameters as the next-best learning-based approach, yet outperforms it by $3.8\%$ in terms of failure detection rate, paving the way toward safely deploying manipulator robots in real-world environments where reliability is non-negotiable.
\end{abstract}



\section{Introduction}
Organizations are increasingly deploying robotic systems in high-stakes environments where failures can cause unwanted delays, damage property, and risk human safety. Among these systems, bimanual manipulators---robots with two coordinated arms---are of special interest because they are designed to perform complex, human-like manipulation tasks such as assembly, tool use, or handling deformable objects. These tasks demand tight coordination between both arms, making the systems especially vulnerable to small perception or control errors. Failures in bimanual manipulation can cascade, compounding damages across fleets and ultimately impeding deployments. To achieve safe deployments, we clearly need scalable methods to reliably detect and mitigate such failures as they arise. 

Failures are typically associated with anomalies, but these can be challenging to define, and often, we are simply interested in flagging any behavior that is unlike a set of observed ``good'' or ``nominal'' behavior. For robots, behavior that meaningfully differs from nominal operation might include unusual sequences of visual or proprioceptive states. However, due to high data-rates and large volumes of data (e.g. modern  robots use multiple 4K camera feeds at $60$Hz for perception), parsing the data in real-time to detect failures is challenging. How can we efficiently represent the high dimensional behavioral sequences of visuomotor robots?





The approach taken in this work is to leverage pretrained foundation vision models to enable the training of world models in a compressed latent space. These world models learn what constitutes good behavior. World models---learned models that forecast future images or other sensory modes conditioned on an action sequence---have emerged as a powerful paradigm in robotics and physical AI. A world model can provide a robot with a means of determining the consequences of potential actions before executing them ~\cite{ward2024optimal,agarwal2025cosmos}, which the robot can use for downstream error detection and recovery, data generation for policy training, and post-failure analysis through counterfactual scenario reasoning. 



We propose a probabilistic variational auto-encoder (VAE) style world model, which provides a quantifiable measure of uncertainty associated with its predictions. We learn only the nominal dynamics---that is, we train the world model entirely on nominal examples---minimizing the uncertainty of predictions while the robot is executing desired behavior. We then use the world model as a runtime monitor, proposing two different error metrics: (i) the intrinsic variance estimated by the VAE at runtime, and (ii) the empirical error obtain by comparing the forecast with the ground truth.  We calibrate failure thresholds for both metrics using conformal prediction \cite{angelopoulos2023conformal}. This enables us to separate anomalous failures from nominal operation at runtime. 


\begin{figure*}[t] 
    \centering
    \includegraphics[width=1.0\textwidth]{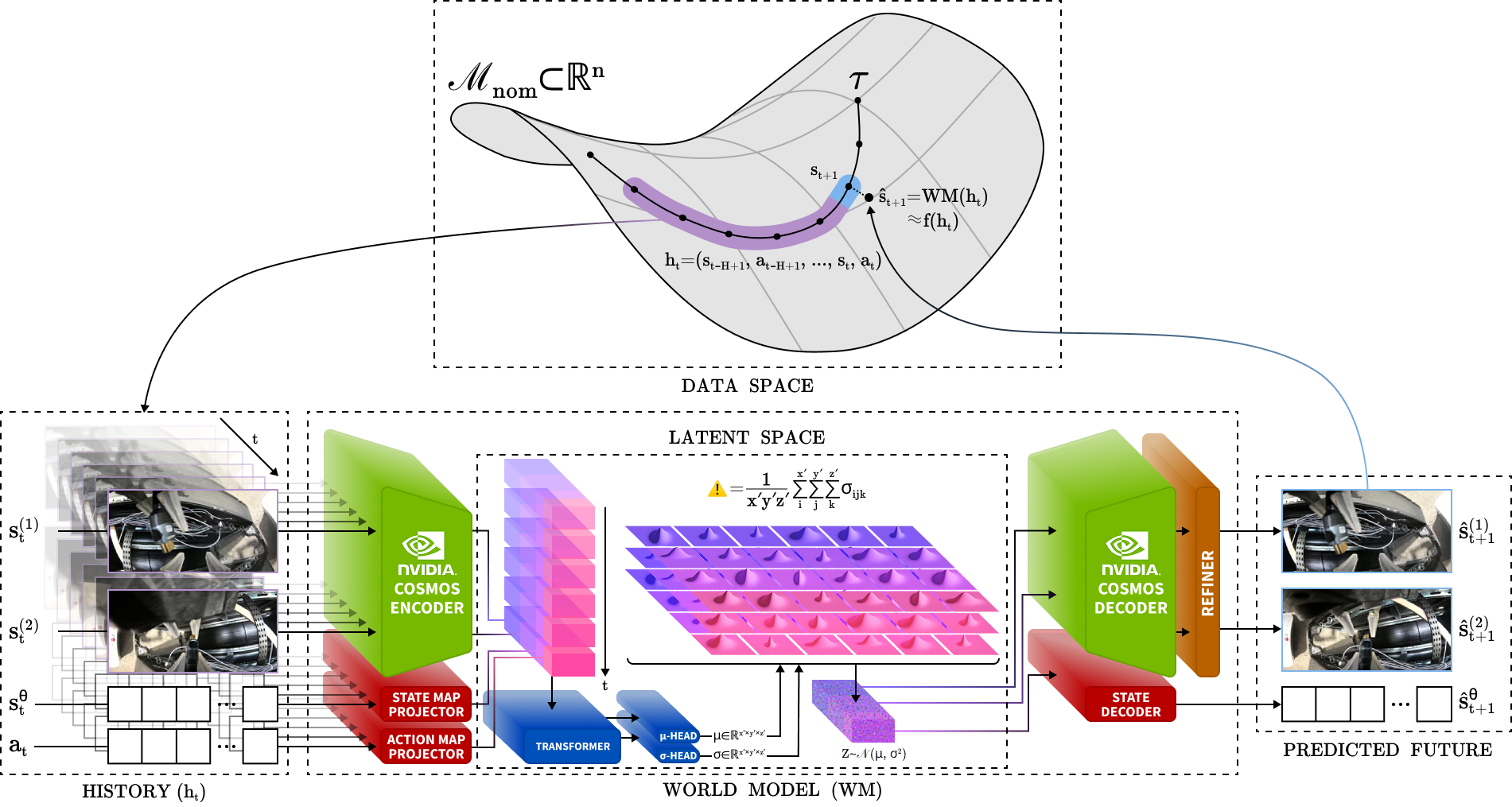}
    \caption{A schematic of the training process for the proposed method. A trajectory $\tau=(s_1, a_1, s_2, a_2, \dots)$ is sampled from the dataset and sliced into training samples consisting of a fixed history window $\mathbf{h}_t = (s_{t-H+1}, a_{t-H+1}, \dots, s_t, a_t)$ and future state $s_{t+1}$. The multi-step, multi-view images captured at the robot's grippers (\sgripith{1}{t}{false} and \sgripith{2}{t}{false}) are encoded using a foundation video encoder (Cosmos) and---alongside learned projections of the historical proprioceptive states \stheta\ and actions $a_t$---are combined into latent feature maps. These are processed sequentially by a transformer to predict a tensor of distributions, the standard deviations of which quantify the uncertainty of the future prediction. This uncertainty is lower for nominal inputs (like those observed during training), and higher for anomalous inputs associated with failures. In other words: $\wm(h_t) \approx f(h_t)$ only when $h_t\in\manifold$. Sampling from this latent distribution gives a latent feature map $Z\sim\mathcal{N}(\mu, \sigma^2)$, which can be decoded and refined into predictions for the next immediate state (\sgripith{1}{t+1}{true}, \sgripith{2}{t+1}{true}, \sthetahat). } 
    \label{fig:hero}
\end{figure*}

Our contributions are thus threefold:
\begin{enumerate}
    \item We propose a probabilistic world model trained within the latent space of NVIDIA's pretrained \textit{Cosmos Tokenizer}\footnote{NVIDIA's Cosmos Tokenizer is a vision autoencoder specialized for manipulator images from the Cosmos video foundation model platform.} \cite{agarwal2025cosmos}. By leveraging this pretrained tokenizer, we create a latent space VAE-based world model with less than $600\text{k}$ trainable parameters.
    \item We propose two methods for failure prediction with our world model: (i) a VAE uncertainty estimate, and (ii) an empirical forecast error, both of which outperform five baseline failure detection methods from the literature.
    \item We introduce the \textit{\datasetname\ dataset}, a new dataset featuring labeled nominal and failure trajectories from real world fleets of bimanual robot manipulators in a data center bring-up and maintenance task.
\end{enumerate}

\section{Related Work}


\input{sections/litrev2}

\section{Problem Formulation}

We consider the problem of identifying trajectories $\tau$ that are considered anomalous. We assume nominal behavior trajectories approximately lie on a lower dimensional manifold $\manifold\subset\realR^{n}$ within the $n$-dimensional data space such that $\dim(\manifold)=m<n$ (see Figure~\ref{fig:hero}). Any $\tau\in\manifold$ that remains close to $\manifold$ is considered nominal, while trajectories that deviate significantly are treated as anomalous failure trajectories.

The system is governed by unknown dynamics $f$
\begin{equation}
    s_{t+1} = f(\mathbf{h}_t),
\end{equation}
where $s_{t+1} \in \realR^n$ is the next state and $\mathbf{h}_t$ is the fixed-length history window of $H$ past state-action pairs at time $t$
\begin{equation}
    \mathbf{h}_t = (s_{t-H+1}, a_{t-H+1}, \dots, s_t, a_t) \in \realR^{H \times (n+d)}.
\end{equation}

Our objective is then to construct a classifier function
\begin{equation}
    C: \realR^{H \times (n+d)} \rightarrow \{0,1\},
\end{equation}
such that
\begin{equation}
    C(\mathbf{h}_t) =
    \begin{cases}
        0, & \text{if } \mathbf{h}_t \text{ is a trajectory near } \manifold, \\
        1, & \text{otherwise},
    \end{cases}
\end{equation}
where $C=0$ indicates nominal behavior and $C=1$ indicates anomalous behavior. 



\section{Methodology}
\label{sec:methods}

Our methodology centers on training a probabilistic, history-conditioned world model (\wm) in the latent space of NVIDIA's {Cosmos Tokenizer} and then using its uncertainty estimates as non-conformity scores within a conformal prediction framework. In addition, we benchmark a set of alternative non-conformity scores adapted from the anomaly detection and  OOD detection literature.

\subsection{Training the World Model}

Following the schematic in Figure~\ref{fig:hero}, the $\wm$'s input is a history of visual observations, proprioceptive states, and actions. Raw images are first embedded using NVIDIA’s Cosmos Tokenizer, yielding latent feature maps that are fused with proprioceptive and action embeddings. A transformer-based sequence model is then trained to predict distributions over future latent feature maps.

\begin{figure*}[t] 
    \centering
    \begin{subfigure}
        \centering
        \vspace{-2mm}
        \includegraphics[width=0.95\textwidth]{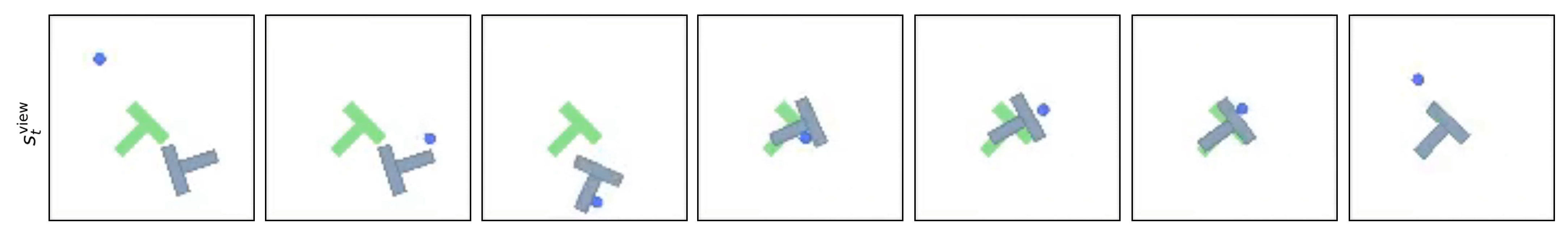}
        \vspace{-2mm}
        \caption*{(a) Each trajectory in the Push-T dataset consists of a $25$fps, $128\times128\times3$ single-view RGB video showing a blue dot (the agent) attempting to push a T-shaped object into a green T-shaped slot. The state $\stheta$ is the $(x,y)$ position of the end effector, and the action $a_t$ (not visualized) specifies the next change in position of the end effector.}
        \label{fig:datapusht}
    \end{subfigure}
    \vspace{5mm}
    \begin{subfigure}
        \centering
        \vspace{-2mm}
        \includegraphics[width=0.95\textwidth]{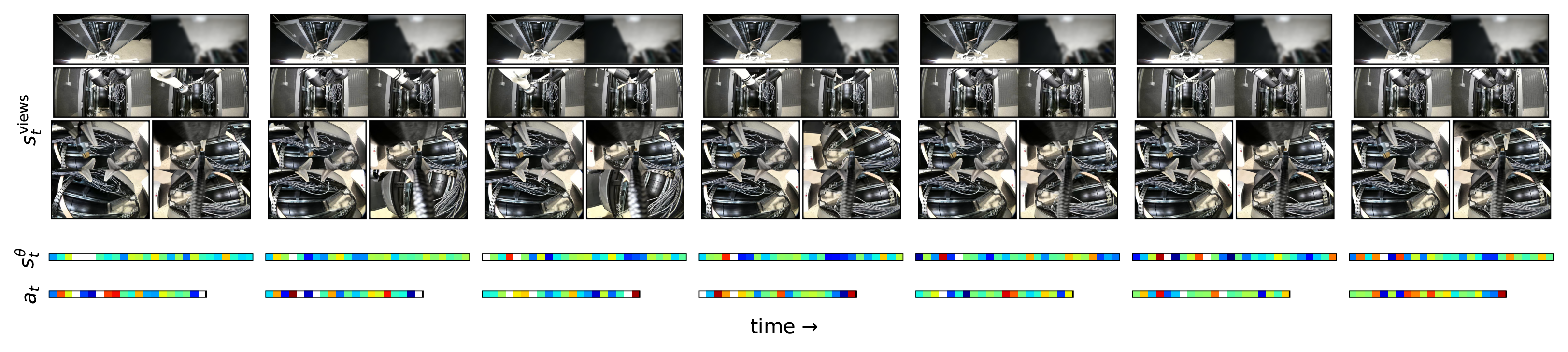}
        \vspace{-2mm}
        \caption*{(b) An illustration of a sequence from a single trajectory from the \datasetname\ dataset. The state comprises $8$ camera views---each $\sim60$fps, $1280\times720\times3$ videos captured from the head (front and back), chest (left and right), left gripper (below and above), and right gripper (below and above) cameras---as well as proprioceptive data at each timestep \stheta$\in\realR^{52}$. Action vectors $a_t\in\realR^{41}$ contain motor torque and gripper activation commands at each timestep. In this work, only the gripper cameras are used as the cable is most visible from these views.} 
        \label{fig:datawatney}
    \end{subfigure}
    
    \caption{Overview of datasets used in this work: (a) Push-T dataset, (b) \datasetname\ dataset.}
    \label{fig:sequence}
    \vspace{-4mm}
\end{figure*}

The \wm\ is trained on ($\textbf{h}_t$, $s_{t+1}$) pairs sampled from only nominal trajectories until the validation loss stops improving. During training, it learns to predict $\wm(\textbf{h}_t)=s_{t+1}$ using a reconstruction loss term that 1) maximizes perceptually accurate reconstructions in pixel space across all camera views (using a perceptual loss $L_{\text{V}}$ \cite{pihlgren2020perceptualloss}) and 2) minimizes the mean square error between the predicted proprioceptive state $\sthetahat$ and the ground truth proprioceptive state $\stheta$)
\begin{equation}
\label{eq:lossinputspace}
\begin{aligned}
\mathcal{L}_{\text{recon}} &=  
    \Big[  
    \frac{1}{N_{\text{views}}} \sum_{i=1}^{N_{\text{views}}}
            \mathcal{L}_{\text{V}}\!\big(s^{(i)}_{t+1},\,\hat{s}^{(i)}_{t+1}\big) 
        \Big]
    +\frac{1}{2}\;\mathcal{L}_{\text{MSE}}\!\big(\stheta,\,\sthetahat\big)
\end{aligned}
\end{equation}
a term that minimizes reconstruction error in the latent space
\begin{equation}
\label{eq:losslatentspace}
\mathcal{L}^{z}_{\text{recon}} = \mathcal{L}_{\text{MSE}}\!\big(z,\,\hat{z}\big)
\end{equation}
a Kullback-Leibler divergence term that ensures that the latent distribution over future states maintains a zero mean and unit standard deviation
\begin{equation}
\mathcal{L}_{\text{KL}} = D_{\text{KL}}\!\Big( \mathcal{N}(\mu, \sigma^2)\;\vert\vert\;\mathcal{N}(0,1) \Big)
\end{equation}
and a negative log likelihood term that ensures that the ground truths would be likely under the learned latent distribution
\begin{equation}
\mathcal{L}_{\text{NLL}} = - \,\mathbb{E}_{z \sim \mathcal{N}(\mu, \sigma^2)} 
    \big[ \log \mathcal{N}(z;\mu,\sigma^2) \big]
\end{equation}
and they are weighted as follows to form the total loss used during training and validation
\begin{equation}
    \mathcal{L} = 
        \sfrac{1}{10} \,\mathcal{L}_{\text{recon}} \;+\;
        2 \,\mathcal{L}^{z}_{\text{recon}} \;+\;
        \sfrac{1}{20} \,\mathcal{L}_{\text{KL}} \;+\;
        \,\mathcal{L}_{\text{NLL}}
\end{equation}


Training begins with a one-step prediction objective, and every $16$ epochs, the autoregressive prediction horizon is doubled as part of a curriculum learning strategy, until a maximum horizon of $32$ steps is reached. This gradual extension of the rollout length stabilizes training and encourages the model to capture longer-term dynamics.




\subsection{Non-conformity scores}
\label{sec:nonconform}

To detect anomalous failures, we need to translate the world model prediction into a non-conformity score. A non-conformity score is a scalar that quantifies how `unlike' the nominal training set the robot's behavior is at a given timestep. Different scoring methods define this {non-conformity} with respect to different properties. We investigate the following non-conformity scores for classifying nominal from failure behavior. 

\subsubsection{\wmunc} defined as the average of each of the standard deviations in the normal distributions over future latents (see Figure~\ref{fig:hero}). For a $3$D tensor of distributions with dimensions $x'\times y'\times z'$ the \wmunc\ score is $\frac{1}{x'y'z'}\Sigma_i^{x'}\Sigma_j^{y'}\Sigma_k^{z'}\sigma_{ijk}$.

\subsubsection{\wmprederr} defined as the discrepancy between the world model’s predicted next-step states and the actual observed states. We compute this in latent space (using Equation~\ref{eq:losslatentspace}), reflecting mismatches in compressed features.


\subsubsection{\logpzo \cite{xu2025can}} refers to the log-probability $- \log p_{\text{flow}}(z_{t+1})$ under a Zero-order (ZO) normalizing flow model trained on nominal latent trajectories. Low log-probability values indicate that an input is unlikely under the nominal distribution.

\subsubsection{\aencrecon\ \cite{japkowicz1995novelty, kirichenko2020why}} the anomaly score is the mean squared error $\| s_t - \hat{s}_t \|_2^2$ between the input $s_t$ and reconstruction $\hat{s}_t$ using an autoencoder trained entirely on nominal states. Failures tend to yield higher reconstruction errors than nominal inputs.

\subsubsection{\aenczsim\ \cite{sinha2024real}}the similarity in latent space between a test embedding $z$ and the nearest embedding from a known safe set of nominal data $z_{\textnormal{safe}}$. Inputs far from the nominal safe set are flagged as anomalous. We use mean-squared error as our similarity function.


\subsubsection{\sparc \cite{balasubramanian2015analysis}} measures the smoothness of a trajectory by looking at the arc length of the Fourier magnitude spectrum of sequential states. A smoother trajectory has less high-frequency content (shorter spectral arc length), while a jerkier trajectory has more high-frequency oscillations (longer arc length). This scalar smoothness measure is used as the non-conformity score.

\subsubsection{\pcakmeans \cite{liu2024multi}} a PCA-based dimensionality reduction followed by K-means clustering into two groups (nominal and failure). The non-conformity score is the distance from the nearest cluster centroid $\min_{c \in \{1,2\}} \| z - \mu_c \|_2$. This is the only method that we benchmark that requires failure data.

\subsubsection{\random} a random baseline that uniformly assigns non-conformity scores at each timestep in the range $[0,1]$ at random, regardless of the information available. This provides a lower bound equivalent to chance-level classification. 


\subsection{Conformal prediction}
\label{sec:confpred}

We use conformal prediction (CP) to calibrate thresholds for each non-conformity score \cite{angelopoulos2021gentle}. For every trajectory, the score sequence is convolved with a uniform triangular filter that most heavily weights the most recent samples, with window length $50$, which has the effect of smoothing out high frequency aberrations. The maximum value is taken as the trajectory-level statistic. 

Thresholds are fit to these trajectory-level statistics using only a held-out set of nominal trajectories, with no access to failure data. At test time, a trajectory is flagged anomalous if its statistic exceeds the $(1-\alpha)$ quantile of the nominal calibration distribution, giving a guaranteed false alarm rate of at most $\alpha$. To reduce bias from how the calibration set is selected, we use a delete-$d$ jackknife with $32$ random permutations, where in each permutation we fit thresholds on half of the held-out nominal trajectories and evaluate on the remaining half, before averaging the resulting thresholds. Table~\ref{tab:classification_results} and Figure~\ref{fig:bar-acc} show these results. 

There are some caveats to using this approach. The validity of these guarantees relies on the assumption that calibration and test data are exchangeable, i.e., the probability of a sequence of values is independent of the order of the sequence. The time series data produced in real-world robotic deployments tends to be temporally correlated and therefore not exchangeable. However, we mitigate this by using a summary statistic per trajectory, meaning that this condition is met if trajectories are exchangeable, even if timesteps within trajectories are not. Furthermore, changing environment conditions, operator behavior, hardware wear, or sensor drift tend to shift the data distribution, so conformal guarantees on the calibration data do not strictly hold for deployment data. In such cases the nominal thresholds may misestimate error rates, highlighting the need for adaptive calibration or re-fitting in long-term deployments.  



\section{Results \& Discussion}



\subsection{The Push-T dataset}

\begin{figure}[htbp]
    \centering
    \includegraphics[width=0.95\linewidth]{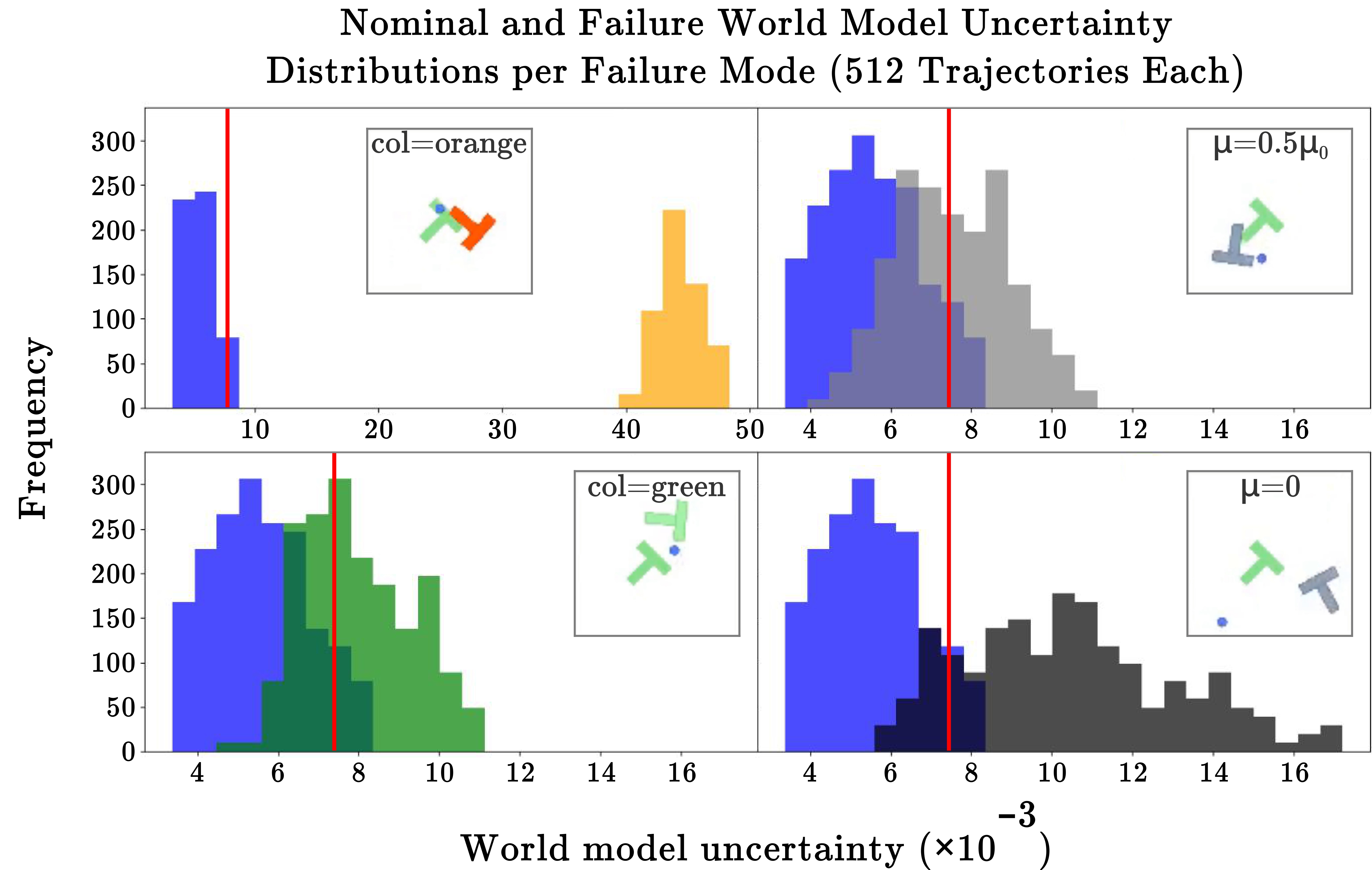}
    \caption{Histograms of the \wmunc\ scores for nominal (blue), anomalous recolored, and anomalous dynamically altered inputs, with overlaid visualizations of what these anomalous inputs look like. In each example the $90\%$ conformal prediction threshold is rendered in red. To test our method's ability to handle visual and proprioceptive anomalies, we alter the color to orange (top left) and green (bottom left) as well as altering the friction to half the nominal friction coefficient (top right) and zero (bottom right).  }
    \label{fig:pushtcombined}
    \vspace{-6mm}
\end{figure}

We use the Push-T environment to create a dataset to test our \wm\ method. The Push-T environment tasks an end-effector agent controlled by a visuomotor diffusion policy \cite{chi2023diffusion} to push a blue T-shaped object into a green T-shaped space (Figure~\ref{fig:sequence}). We generate $1028$ nominal rollouts for training, and $128$ nominal rollouts for validation (each with different initialization seeds). We generate a further $512$ nominal and $2048$ failure rollouts ($512$ each across $4$ distinct failure modes) for evaluating our methods. The four modes each have a unique disturbance which we apply to force failure: changing the T from blue to orange, changing the T from blue to green, reducing the friction of the T, and removing friction altogether.

As outlined in Figure~\ref{fig:pushtcombined}, we find that the \wmunc\ scores for the OOD inputs are generally higher than the scores for the nominal trajectories across multiple OOD modes. In particular, we find that our method---which ingests both visual and kinematic data---is able to detect visual anomalies (changing colors) and changes to the dynamics (altered friction), clearly separating nominal and anomalous behavior.

\subsection{The \datasetname\ dataset}
\label{sec:dataset}


We introduce the \datasetname\ dataset to further test our approach. The dataset features $83$ nominal trajectories for training and validation, $\numnominalfit$ held-out nominal trajectories for fitting conformal prediction thresholds, $\numnominaleval$ held-out nominal trajectories for evaluation, and $\numfailure$ held-out failure trajectories for evaluation. The primary failure mode in this dataset is the cable being dropped by the robot's manipulators as it attempts to plug in the cable (as illustrated in Figure~\ref{fig:detect}).  

\begin{figure}[h]
    \centering
    \fcolorbox{white}{white}{\includegraphics[width=0.32\linewidth]{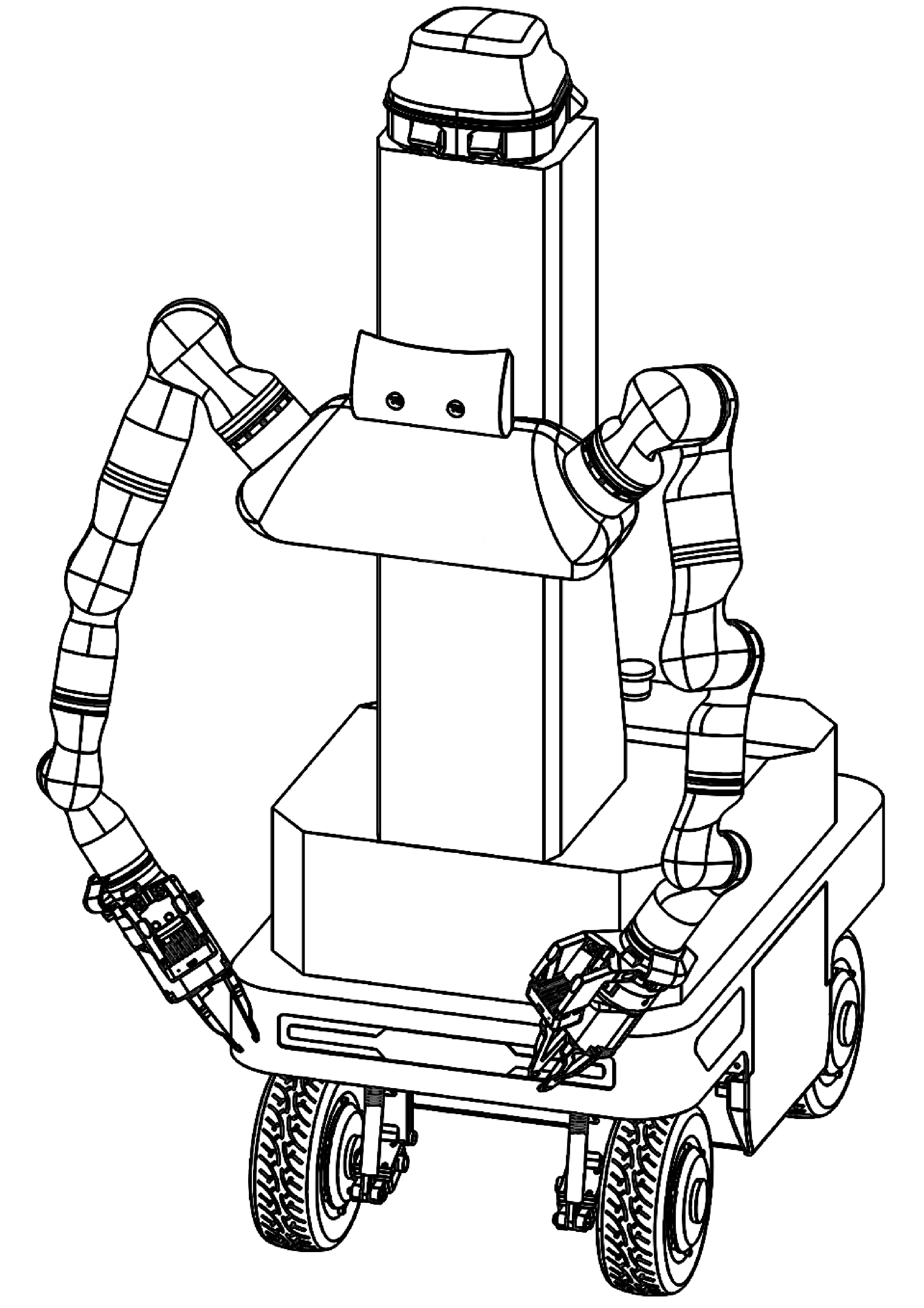}}
    \caption{The WR1 Data Center Mobile Manipulator robot.}
    \label{fig:wr1robot}
    \vspace{-2mm}
\end{figure}

The dataset was collected by a WR1 robot (see Figure~\ref{fig:wr1robot}) in a data center environment operating with a manual policy steered by a trained data center technician via a secured remote link from over 7000 miles away. The robots operate using a segregated private network link and proprietary secured communication protocol enabling ultra low latency communication for remote inference and operation. 


Each trajectory is between $18$ seconds and $3$ minutes long, and features four synchronized $\sim60$fps, $1280\times720\times3$ RGB video feeds from cameras mounted on the robot at head height, chest height, and on the end of each manipulator (as illustrated in Figure~\ref{fig:sequence}). Moreover, each state also features a $52$-dimensional proprioceptive state vector and $41$-dimensional action vector. 

In Table~\ref{tab:classification_results} (and Figure~\ref{fig:bar-acc}), we benchmark each of the methods described in Section~\ref{sec:nonconform} on the \datasetname\ dataset using the method described in Section~\ref{sec:confpred} to fit classification thresholds. We find that our methods consistently outperform competing learning-based methods despite having only $\sim\sfrac{1}{20}^{th}$ of learnable parameters ($569.7$k compared to approximately $10$M) due to being trained in the latent space of a video foundation model. Moreover, our methods categorically outperform statistical methods in terms of classification accuracy. 

\begin{table}[!htbp]
    \centering
    \renewcommand{\arraystretch}{1.2}
    \setlength{\tabcolsep}{6pt}  
    \captionsetup{justification=raggedright}  
    \caption{\normalfont Classification performance of different methods on the \datasetname\ dataset, using an $85\%$ conformal prediction threshold.
    }
    
    \begin{tabular}{l c c c}
        \toprule
        & \multicolumn{3}{c}{Avg. Classification Accuracy ($32$ folds, \%)} \\
        \cmidrule(lr){2-4}
        & Nominal & Failure & Weighted total \\
        Method & (\symbolnominaleval$=$\numnominaleval) & (\symbolfailure$=$\numfailure) & (\symboltotal$=$\numtotal) \\
        \midrule
        \wm\ uncertainty \textbf{(ours)}        & $87.9\pm17.0$ & $\mathbf{95.1\pm5.5}$ & $\mathbf{92.0\pm6.4}$ \\
        \wm\ pred. error \textbf{(ours)}        & $\mathbf{88.3\pm17.8}$ & $87.5\pm12.3$ & $87.9\pm6.4$ \\
        \logpzo                                 & $86.8\pm20.4$ & $91.3\pm6.7$ & $89.3\pm6.8$ \\
        \aenc\ recon. error                     & $80.6\pm20.5$ & $45.8\pm18.6$ & $61.0\pm4.2$ \\
        \aenczsim\                              & $80.7\pm20.3$ & $55.2\pm21.4$ & $66.4\pm6.1$ \\
        \sparc\                                 & $64.7\pm35.6$ & $25.3\pm18.1$ & $42.6\pm6.8$ \\
        \pcakmeans\                             & $66.9\pm33.7$ & $34.4\pm8.9$ & $48.6\pm12.6$ \\
        \random\                                & $55.3\pm34.9$ & $25.7\pm20.5$ & $38.7\pm6.4$ \\
        \bottomrule
    \end{tabular}
    \label{tab:classification_results}
\end{table}

Interestingly, we find that the empirical coverage of our conformal prediction bounds in the evaluation set does not always match the presumed coverage set by the $(1-\alpha)$ quantile ($85\%$ in practice), implying that the evaluation data and fitting data do not necessarily come from the same underlying data distribution (see Section~\ref{sec:confpred} for further discussion of the assumptions relating to conformal prediction). We expect that this effect is due to the relatively small size of the total nominal set; it is possible to repeatedly select a biased fitting subset of the total nominal set that is meaningfully different from the remaining evaluation subset, making it possible to not observe an $\sim85\%$ nominal classification accuracy. 

Note that \random\ in this case does not mean randomly guessing the class, but randomly assigning a non-conformity score to each timestep of the trajectory, and then applying conformal prediction to a class-imbalanced evaluation dataset, hence the $<50\%$ class-weighted accuracy on a binary classification task.

\subsection{The \wmunc\ score is correlated with approaching or ongoing anomalous failures }

\begin{figure*}[!t]
    \centering
    \includegraphics[width=\textwidth]{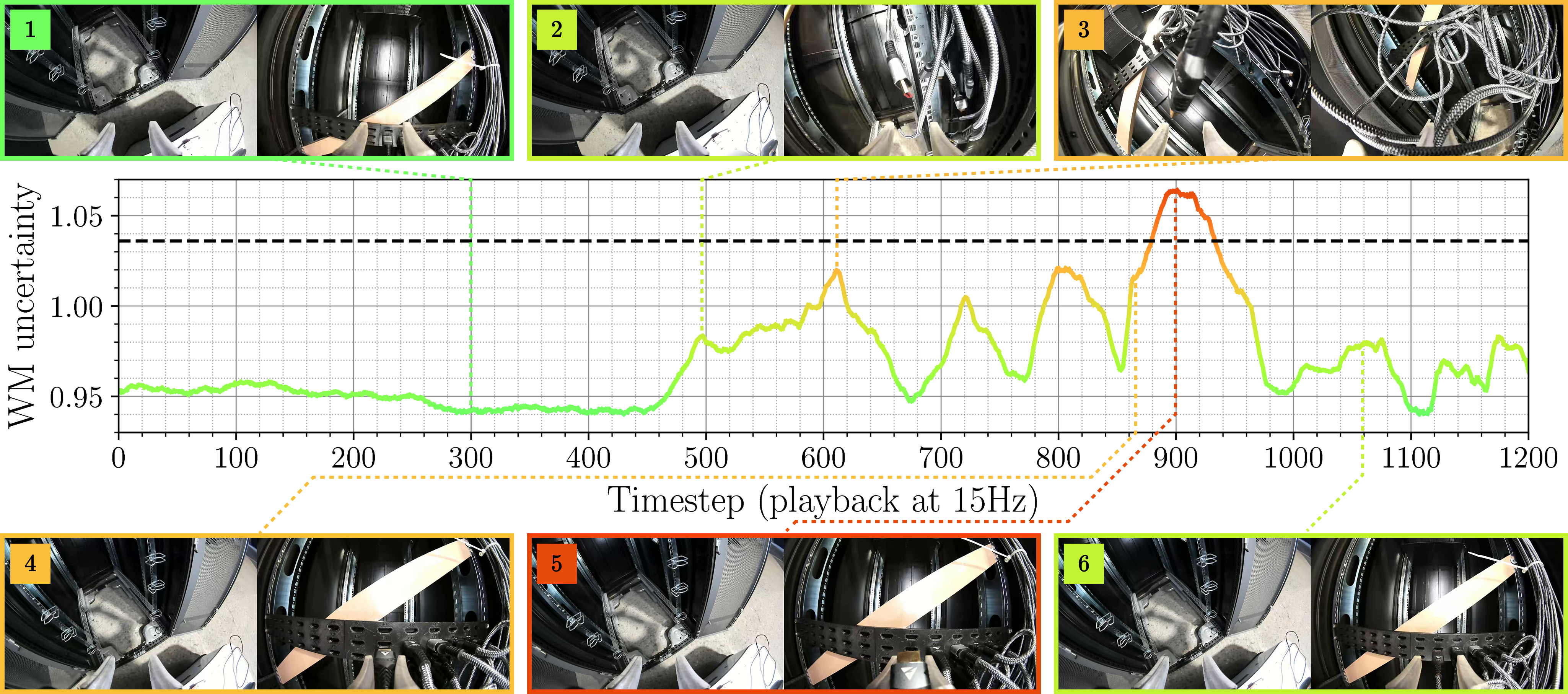} 
    \caption{An example failure detection on an unseen failure case, visualizing captures from the left and right gripper cameras. (1) Nominal behavior---the robot is not gripping or attempting to manipulate a cable. (2) The uncertainty rises as the gripper approaches a cable and begins a sequence of grab attempts. Uncertainty tends to increase when a cable head is in view. (3) The first of three aborted grab attempts, such periods are usually associated with increased uncertainty as grab attempts represent a small portion of the nominal training set. Note that these do not constitute failures. (4) Uncertainty increases rapidly as the robot focuses instead on unplugging a cable that is already plugged in (observe the force being applied onto the cable head by the compliant grippers). (5) An actual failure occurs (dropped cable) as the robot fumbles the cable head during the unplug maneuver. (6) With the dropped cable fully out of frame, the robot is again not gripping or attempting to manipulate a cable as in (1). Uncertainty again reduces to nominal levels. Note that the world model is trained on proprioceptive state data, and is action conditioned, so some part of the predicted uncertainty likely stems from proprioceptive features not visualized in this figure.
    }
    \label{fig:detect}
\end{figure*}

Figure~\ref{fig:detect} outlines an unseen failure case where the \wmunc\ metric clearly increases and decreases as inputs become more or less correlated with moments of impending or present failures. Moreover, although we only classify two regimes; nominal and failure, we do observe distinct regions with different amounts of relative `safety' for the robot. In particular, we observe that periods where the robot is not holding a cable---and it is thus impossible to drop the cable (failure)---are associated with the lowest \wmunc. Moreover, the score rapidly increases prior to the cable being dropped, despite the cable still visibly being grasped by the grippers. We hypothesize that this is due to the world model correlating certain proprioceptive state/action sequences with nominal behavior, and not observing such inputs in these pre-failure timesteps.

\subsection{\wmunc\ is a better predictor of anomaly than \wmprederr}

\begin{figure}[h]
    \centering
    \vspace{-5mm}
    \includegraphics[width=0.9\columnwidth]{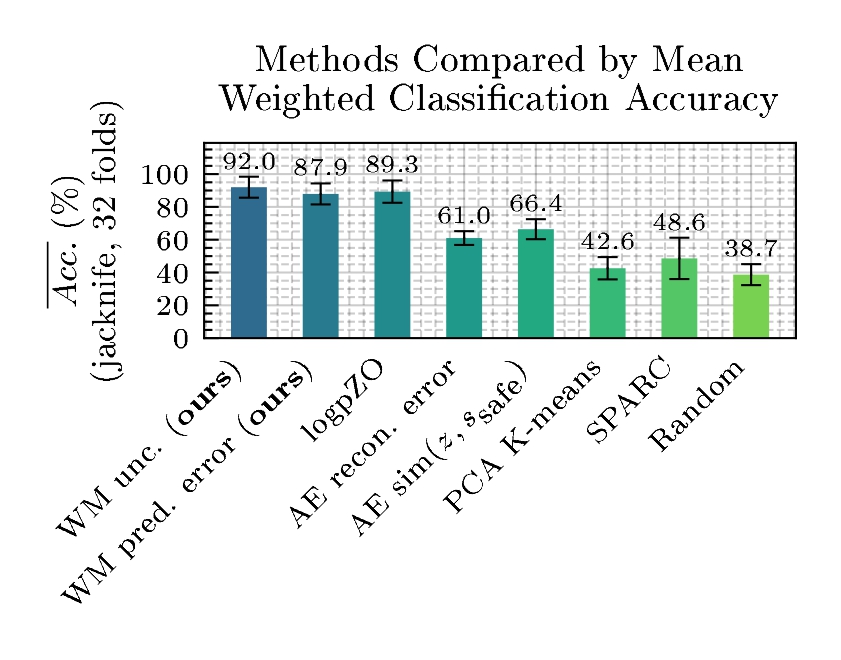}
    \vspace{-2mm}
    \caption[Caption for figure]{We find that our \wm-based approaches outperform competing methods on the \datasetname\ dataset, presumably because of how they uniquely incorporate historical context. See Table~\ref{tab:classification_results} for a class-based breakdown.}
    \label{fig:bar-acc}
    \vspace{-2mm}
\end{figure}


Figure~\ref{fig:bar-acc} illustrates that our \wm\ methods outperform competing methods on the \datasetname\ dataset, though we note that \wmunc\ is a more reliable non-conformity score than \wmprederr. We expect that this is due to the prediction error on a single outcome being less informative than the model’s own distributional uncertainty, since a low-error match can occur by chance even when the input is off-manifold, whereas high predictive variance more reliably implies anomalous inputs.


\subsection{Learning-based methods are slower but can still be executed in real-time}

\begin{figure}[h]
    \centering
    \vspace{-0mm}
    \includegraphics[width=0.9\columnwidth]{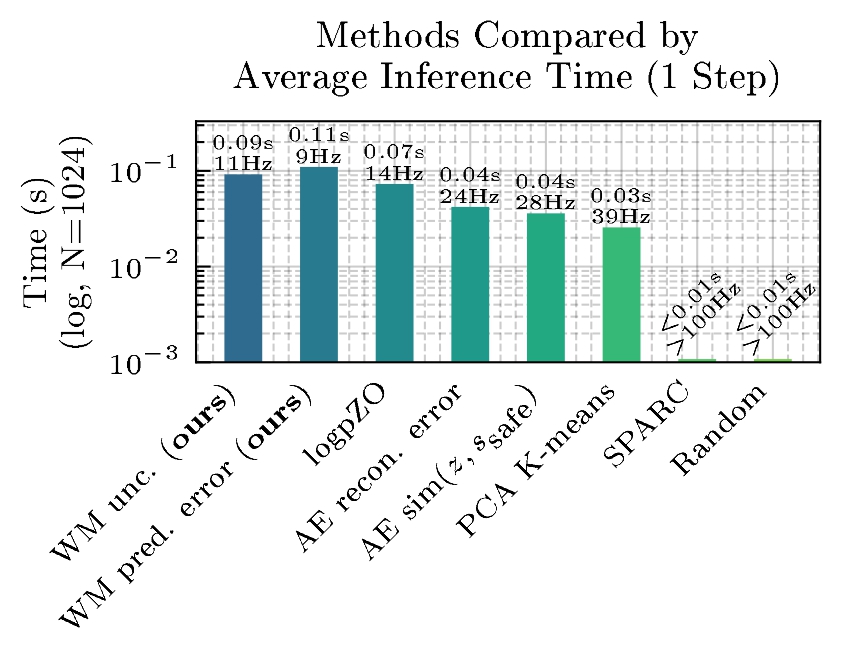}
    \vspace{-2mm}
    \caption{The time taken to generate a scalar non-conformity score value for a single step, for each method. 
    }
    \label{fig:bar-time}
    \vspace{-4mm}
\end{figure}



We benchmark the runtime of each method by measuring the wall-clock time to compute a single scalar non-conformity score. As illustrated in Figure~\ref{fig:bar-time}, all methods operate comfortably above $9$Hz, satisfying the threshold for real-time execution in our robotic setting. Statistical baselines such as \sparc\ and \pcakmeans\ are the fastest, while deep learning-based methods are slower due to the forward passes through neural architectures. Among the learning-based approaches, scores that operate directly in latent space (\aenczsim) are faster than their counterparts requiring reconstruction to pixel space (\aencrecon), as decoding latent codes requires further processing. Although \wm-based methods are the slowest overall, their throughput remains sufficient for deployment, and world models do provide ancillary utility in a robotics context (e.g. enabling planning).

\section{Conclusion}

We present a general method for detecting anomalous failures using a probabilistic world model trained entirely on data featuring nominal task completions. We find that world model uncertainty is a reliable indicator of anomalous behavior, effectively distinguishing nominal from failure trajectories in real-time, in a real-world, bimanual robot manipulator setting.


\textit{Limitations:} Our approach uses conformal prediction, but violates formal assumptions on exchangeability. Nonetheless, our results show empirically that the conformal calibration is effective. Our method may flag false errors due to benign distribution shifts (e.g., background color changes). The method also inherits currently unknown biases from the pretrained tokenizer. We expect that these limitations could be resolved with more data, and by testing the efficacy of other pretrained encoders as they become available.


\textit{Future Work:} Promising avenues for future work include exploring longer and more compact historical representations to improve detection of longer-term task-progression failures, conducting feature importance analysis to quantify the contribution of proprioceptive and visual data to the \wm\ uncertainty metric, and finally, extending the approach to fully autonomous manipulation policies. In particular, leveraging the \wm\ for simultaneous failure detection \textit{and correction} using optimization procedures that solve for the action sequence that minimizes \wm\ uncertainty may provide a solution that neatly improves the robustness of bimanual manipulators. 

\vspace{-4mm}



\section*{Acknowledgments}

The authors thank Watney Robotics for providing funds, access to robot hardware, and datasets which assisted the authors with their research.

The authors thank Toyota Research Institute (TRI) for providing funds to assist the authors with their research, but this article solely reflects the opinions and conclusions of its authors and not TRI or any other Toyota entity.

Generative AI tools and technologies were used in this work to identify related literature (ChatGPT, Perplexity AI), sense-check concepts (ChatGPT), and tab-autocomplete code (GitHub Copilot running in Visual Studio Code). 


\bibliographystyle{IEEEtran}
\bibliography{refs}


\appendix
\label{sec:reprod}







All experiments were executed on a \texttt{gpu\_1x\_h100\_pcie} instance on Lambda Cloud. The system was equipped with a single NVIDIA H100 PCIe GPU with 80\,GB of VRAM (driver version 570.158.01, CUDA~12.8), paired with an Intel Xeon Platinum 8480+ CPU. 




        
        


        

        

\end{document}

%% file: sections/litrev2.tex
\subsection{Classical and Statistical Methods}


Classical methods, including control charts, hypothesis testing, residual analysis, and change-point detection, are widely used across disciplines such as quality control and computer networks \cite{hastie_elements_2009} to detect anomalies that may result in failure. They are simple in formulation and benefit from interpretability and minimal computational overhead \cite{chandola2009anomaly}, but lack the representational power and adaptability of modern learning-based techniques. Moreover, these methods are ill-suited for robotics, where strong assumptions such as stationarity, noise independence, and accurate dynamics models often break down \cite{ho2022towards}. Robotic systems are high-dimensional, multimodal, and context-dependent, with temporal correlations that classical techniques fail to capture. 

Ensemble-based statistical methods attempt to address some of these shortcomings by aggregating multiple detectors, each sensitive to different statistical properties of the data. Common ensemble approaches for OOD detection include ensembles of classifiers \cite{contreras2024safe} or generative models \cite{vyas2018out}, and hybrid methods that incorporate Bayesian uncertainty estimation \cite{blundell2015weight, liu2020simple}. These methods estimate uncertainty from limited knowledge of the data distribution, flagging high variance or predictive disagreement as anomalies. 

\subsection{Autoencoder and Embedding-Based Methods}

Representation learning methods aim to discover feature spaces in which nominal and anomalous states can be more easily separated. Autoencoder-based approaches learn these feature spaces via self-supervised reconstruction, with reconstruction error serving as an anomaly score \cite{richter2017safe, oza2019c2ae}. 


Similarly, embedding-based methods operate on the assumption that in-distribution data lies in compact regions of feature space, while anomalies lie farther from the nominal manifold \cite{chandola2009anomaly}. Non-conformity scores are then defined in terms of distance or similarity between a test embedding and nominal embeddings. Parametric approaches include Mahalanobis distance, which assumes Gaussian structure in the embedding space \cite{lee2018simple}, while non-parametric approaches rely on neighborhood similarity \cite{bergman2020classification-based, tack2020csi}. Embedding-based approaches are model-agnostic, and integrate naturally with supervised or self-supervised representation learning. Despite their flexibility, embedding-based methods depend heavily on the quality of the learned feature space, limiting reliability under distribution shifts.  



\subsection{Distribution Models}

Distribution modeling approaches explicitly attempt to learn the nominal data distribution, with the assumption that out-of-distribution (OOD) samples will have low likelihood under the model. Variational autoencoders \cite{xiao2020likelihood}, autoregressive models \cite{wang2020further}, and normalizing flows \cite{ward2025improving,xu2025can} have all been applied to this problem. However, it has been shown that such models can assign high nominal likelihoods to anomalous data \cite{nalisnick2019deep}, particularly in the case of normalizing flows, which tend to capture low-level pixel correlations rather than high-level semantic features \cite{kirichenko2020why}. These methods are also computationally demanding, requiring large amounts of diverse nominal data for training and significant resources for inference.

\subsection{Modern methods}

Some recent works have proposed to use world models, such as DINO \cite{zhou2024dino} for error recovery \cite{nakamura2025generalizing} and human intent alignment \cite{wu2025foresight} in visuomotor robot policies. Generally speaking, world models learn from a training dataset how a robot's state evolves given a sequence of actions. When test-time inputs deviate from the training dataset, changes in the world model's confidence or epistemic uncertainty allow for OOD detection \cite{ho2026world, seo2025uncertainty}. 


Finally, large language model (LLM)-based approaches transform visual or sensory inputs into symbolic or natural language descriptions, which are then evaluated for plausibility by an LLM \cite{elhafsi2023semantic, sinha2024real}. At a semantic level, this mirrors human anomaly detection, where context and meaning guide judgments rather than low-level patterns. Such approaches promise better generalization to novel scenarios, interpretability through natural language explanations, and the integration of common-sense reasoning. However, challenges remain, including sensitivity to prompt design, latency, and susceptibility to hallucination or bias. While still nascent, these methods highlight the potential of foundation models to reason about anomalous behavior in robotics.